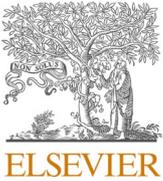
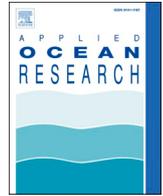

Research paper

# Defect Detection in Synthetic Fibre Ropes using *Detectron2* Framework

Anju Rani [*], Daniel Ortiz-Arroyo, Petar Durdevic

*Department of Energy, Aalborg University, Niels Bohrs Vej 8, Esbjerg, 6700, Denmark*

ABSTRACT

Fibre ropes with the latest technology have emerged as an appealing alternative to steel ropes for offshore industries due to their lightweight and high tensile strength. At the same time, frequent inspection of these ropes is essential to ensure the proper functioning and safety of the entire system. The development of deep learning (DL) models in condition monitoring (CM) applications offers a simpler and more effective approach for defect detection in synthetic fibre ropes (SFRs). The present paper investigates the performance of Detectron2, a state-of-the-art library for defect detection and instance segmentation. Detectron2 with Mask R-CNN architecture is used for segmenting defects in SFRs. Mask R-CNN with various backbone configurations has been trained and tested on an experimentally obtained dataset comprising 1,803 high-dimensional images containing seven damage classes (placking high, placking medium, placking low, compression, core out, chafing, and normal respectively) for SFRs. By leveraging the capabilities of Detectron2, this study aims to develop an automated and efficient method for detecting defects in SFRs, enhancing the inspection process, and ensuring the safety of the fibre ropes.

## 1. Introduction

In offshore industries, SFRs made from materials like Dyneema are nowadays used by cranes to lift and hoist heavy equipment to the platform facilities. These newly developed ropes are lightweight, hydrophobic, UV resistant, and offer 15 times higher strength than the traditional steel wire ropes (SWRs) (McKenna et al., 2004, Hoppe, 1997). SWRs must consider their weight when determining the maximum depth for payload deployment. Conversely, SFRs like Dyneema are naturally more buoyant than steel, allowing payloads to be deployed at greater depths using smaller cranes. This has led to a rising trend among companies in developing fibre rope cranes customized for industrial use. However, due to their prolonged use in critical systems, SFRs suffer from plastic wear, abrasive wear, slack strands, and slack wires, necessitating continuous testing and evaluation of defects and damages to estimate their remaining useful life (RUL) (Feyrer, 2007; Ridge et al., 2001, Oland et al., 2017, Onur and İmrak, 2011). Also, SFRs present unique challenges for defect detection due to its non-linear behavior and susceptible to diverse range of damages during its lifetime such as chafing (abrasion/wear), core out, plackings (loops formation due to strands out), internal breakages, etc., making their inspection challenging. Manual inspection of such SFRs is a costly, challenging, inefficient, and time-consuming task. Therefore, there is a need to develop a faster and more automatic inspection scheme for inspecting damages in SFRs, reducing the total maintenance cost. In the literature, non-destructive testing (NDT) techniques (Onur and İmrak, 2011) such as magnetic (Antin et al., 2019, Yan et al., 2017), acoustic emission (AE) (Zhang et al., 2006), ultrasound, *X*-rays, and *γ*-rays (Casey and Taylor, 1985), fibre optics (Huang et al., 2020, Paixao et al., 2021), and computer vision method (CVM) (Vallan and Molinari, 2009, Platzer et al., 2009) etc. have been used for defect detection in SWRs (Oland et al., 2017). (Falconer et al., 2017) utilized CVM to monitor the length and width measurements of SFRs under tension testing as a condition indicator. Later, (Falconer et al., 2020) combined CVM and thermal monitoring during cyclic bend over sheave tests for CM of fibre ropes. (Weller et al., 2015) conducted a series of experiments involving a mixture of load/unload, harmonic, and steady load tests on SFRs to quantify their stiffness and damping properties. (Weller et al., 2015) investigated the influence of load on the condition of the rope as a performance measurement in the mooring system. (Halabi et al., 2023) conducted experiments to investigate the tensile characteristics of SFRs. Subsequently, (Halabi et al., 2023) utilized artificial neural network (ANN) models to predict the tri-linear stress–strain profiles of the ropes under investigation.

In recent years, DL models have been used for various applications such as condition monitoring, image analysis, video surveillance, etc. DL models process data collected from sensors and cameras to detect defects or damages. DL models in computer vision for object detection have been designed to provide additional information on the location and shape of objects. In condition monitoring applications, these models have shown remarkable performance in detecting damages for a variety of settings such as electrical systems (Tabernik et al., 2020), conveyor belts (Yang et al., 2019), manufacturing industries (Yang et al., 2020), railway tracks (Wei et al., 2019), roads (Pham et al., 2020) etc. Jalonen et al. (Jalonen et al., 2023) conducted experiments to collect an image dataset consisting of normal, worn, and damaged SFRs. (Jalonen et al., 2023) designed a CNN-based model to detect damages in SFRs. The results indicated that CNN9 and CNN15 outperformed other presented models for detecting damages. Recently developed one-stage detectors such as YOLO version (v2, v3, v4, v5, v6, v7, and v8) (Zhu et al., 2021) and single shot multi-box detector (SSD) (Liu et al., 2016) have been used in defect detection due to their simple structure and fast speed. The single-stage detector utilizes the same feed-forward network fully convolutional network (FCN) (Long et al., 2015) for detecting bounding boxes (BBox's) for object classification. FCN architecture allows the model to efficiently process the input image to generate predictions for BBox's coordinates and corresponding object classes. This integrated approach simplifies the detection pipeline and reduces computational complexity, making it suitable for real-time applications. However, its accuracy is comparatively lower than two-stage detectors based on region-based convolutional neural networks (R-CNN) (Bharati and Pramanik, 2020). The two-stage detectors have separate stages for region proposal and classification, allowing more precise localization and improved classification accuracy.

R-CNN was a breakthrough for object detection, and semantic segmentation (Cai and Vasconcelos, 2018, Wang et al., 2017), as it introduced a novel approach by combining DL with a region proposal network (RPN) generating object regions in an image. These proposed regions are then processed by a CNN-based classifier to determine the presence of objects and classify them into predefined categories. This multi-stage approach improved accuracy and efficiency compared to traditional methods. (Ye et al., 2024) compared several semantic segmentation models including U-Net (Ronneberger et al., 2015), PSP-Net (Zhao et al., 2017), DeepLabv3+ (Chen et al., 2018), and HR-Netv2 (Sun et al., 2019) for detecting faults in a video dataset of SFRs. Variations such as Fast R-CNN (Girshick, 2015), Faster R-CNN (Ren et al., 2015), and FCN (Long et al., 2015) have demonstrated the influence of deep R-CNN in achieving high performance for semantic segmentation tasks to detect and localize objects in an image. Semantic segmentation classifies each pixel in an image into a set of pre-defined classes. Later, Mask R-CNN (He et al., 2017, Bolya et al., 2019) framework was developed, enabling instance segmentation that not only classifies each pixel as in the case of semantic segmentation but divides an image into separate, distinct parts corresponding to the object of interest. Panoptic segmentation (Kirillov et al., 2019) combines the features of semantic and instance segmentation for a better understanding of an image by providing class labels for each pixel and unique instance IDs for each object. Recently developed Detectron2, a successor of the Detectron is a state-of-the-art library providing a wide range of functionalities for object detection and segmentation tasks (Wu et al., 2019). Detectron2 includes a collection of models, such as Mask R-CNN, Faster R-CNN, Fast R-CNN, RetinaNet, TridentNet, DensePose, Cascade R-CNN, and Tensor-Mask, etc. Detectron2 has support for three different types of segmentation: semantic, instance, and panoptic segmentation, respectively. In (Wen et al., 2021), Detectron2 with Mask R-CNN architecture was used to detect pores and cracks in scanning electron microscope (SEM) images of metallic additive manufacturing (AM) parts. The study accurately identifying over 90% of the defects present in the testing images. (Yagüe et al., 2022) identified defects of irregular and complex boundaries with high precision on X-ray images obtained from automotive parts using Detectron2 with Faster R-CNN architecture. (Ali et al., 2022) utilized Detectron2 with Faster R-CNN to detect COVID-19 from chest X-ray images. The experiment was conducted with different baseline models to assist radiologist while making critical decisions. It is important to emphasize that machine learning methods, including deep CNNs and R-CNNs, have been utilized for condition monitoring of SWRs in previous studies. Furthermore, object detection models like Fast R-CNN, Faster R-CNN, Mask R-CNN, Cascade R-CNN, and YOLO have been effectively used for defect detection across a various material, such as metals, composites, bridges, etc. This paper aims to bridge the existing methodological gaps to demonstrates promising findings in detecting defects in fiber ropes.

In the present paper, the state-of-the-art Detectron2 framework with the Mask R-CNN architecture was used to detect defects on experimentally collected high-resolution SFR image datasets. Detectron2 utilizes pre-trained models to harness the power of transfer learning (TL). These models are pre-trained on massive image datasets and then fine-tuned on the specific task. In this way, Detectron2 is a scalable and efficient approach for faster inference on real-time scenarios. The Mask R-CNN architecture in Detectron2 effectively handles the complex and overlapping damage instances thereby ensuring accurate and detailed segmentation. This capability enables automated analysis of SFRs with high accuracy, diminishing reliance on manual inspection and minimizing the potential for human error. The primary objective of this study is to train a model capable of accurately detecting and segmenting the defects in the SFRs dataset by assigning specific labels to each pixel or region of interest. The model's performance is then assessed using appropriate evaluation metrics. The novelty and difficulty of this work lie in the following aspects:

1. Defect detection in SFRs is a challenging task due to the complex and non-linear behavior of ropes, as well as the diverse range of damages they can sustain during their lifetime, such as chafing, plackings, core damage, and compressions.
2. This work addresses the limitations of manual inspection of SFRs, which is costly, time-consuming, and prone to human error, by introducing an automated and efficient DL-based approach for defect detection.
3. The study aims to modernize traditional inspection practices in the offshore industry by offering a more precise, efficient, and automated solution for identifying defects in SFRs, ultimately enhancing industrial safety and operational efficiency.
4. The study utilizes an experimentally obtained high-dimensional image dataset comprising 1,803 SFRs with seven different types of defect class collected from three different ropes, making it a realistic and challenging dataset for defect detection.
5. Detectron2 with Mask R-CNN architecture has been utilized for defect detection, even when they overlap or occlude each other, by providing instance-level segmentation masks.

The rest of the paper is structured as follows: Section 2 provides an overview of the proposed Detectron2 framework, covering aspects such as experimental setup, data collection methodology, annotation process, and model configuration. Additionally, the evaluation metrics employed in this study are presented, which serve as the basis for assessing the model's performance. Section 3 discusses the results, showcasing the performance of the proposed work. Finally, Section 4 concludes the paper, summarizing the findings and discussing their implications.

**2. Model**

Fig. 1 illustrates the fundamental architecture of the Detectron2 framework. The architecture consists of two main stages: the backbone and the head. The backbone stage allows a combination of sub-networks, including ResNet, ResNeXt, feature pyramid networks (FPN), and other similar architectures, as long as their input and output dimensions are compatible. Each network has its own strengths and characteristics making them suitable for specific tasks. ResNet introduces residual blocks with skip connections addressing the challenge of training very deep neural networks allowing them to learn complex features with high accuracy. ResNeXt, an extension of the ResNet network introduces cardinality (number of parallel internal paths or branches) within a single block to enhance its feature extraction capabilities. This improves the network accuracy in comparison to increasing the number of layers of the network. On the contrary, FPN addresses the issue of multi-scale object detection and feature learning by creating a pyramid of features making it effective for tasks where objects can appear in various sizes. In the present work, the performance of different backbone architectures is compared to choose the most suitable architecture for detecting defects in SFRs. The chosen architecture is utilized to extract significant features from the input image. These extracted features are then directed to a region proposal network (RPN) responsible for determining the presence of objects within specific regions. Subsequently, these regions are processed by the head stage through an FCN layer to predict the coordinates of BBox's and associated class labels. During this phase, Intersection over Union (IoU) metrics are computed. The correct predictions of BBox's can be calculated using the IoU score defined by:

$$IoU = \left[\frac{area(PB \cap GB)}{area(PB \cup GB)}\right] \quad (1)$$

where IoU $\geq$ 0.5 depicts a good match while no match otherwise. Also, $\cap$ and $\cup$ represents intersection and union between predicted BBox's (PB) and actual/ground-truth BBox's (GB) respectively.

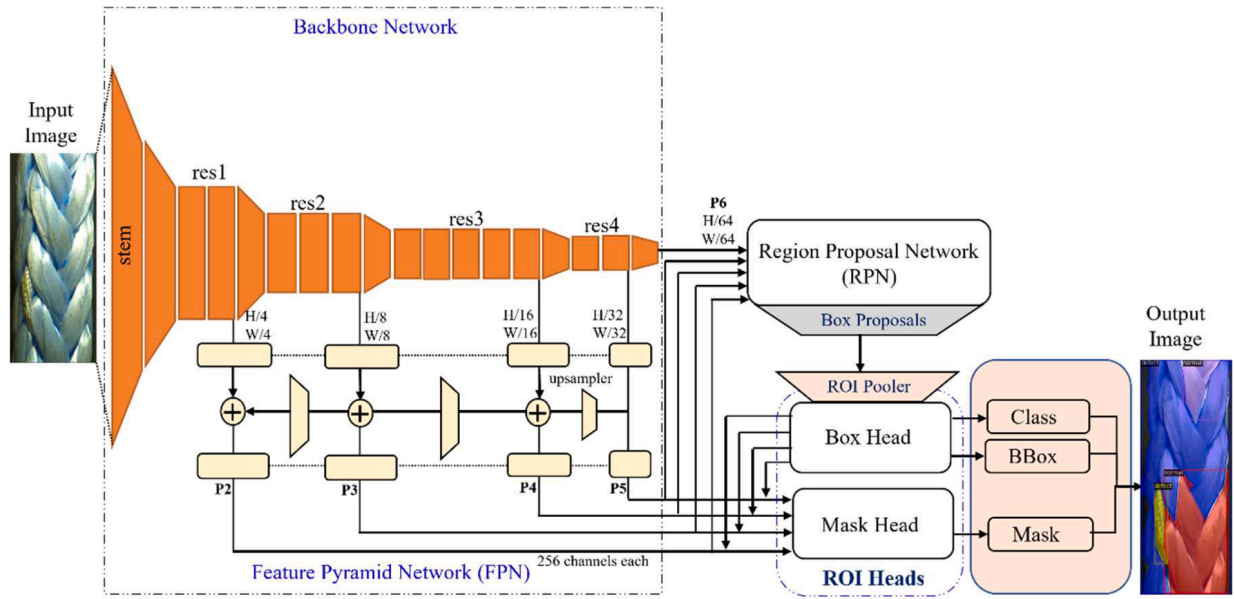

**Fig. 1.** Architecture of Detectron2 model.

The capabilities of the model are further enhanced by incorporating a mask branch into the existing architecture. This introduces a segmentation mask for each detected object region, facilitating precise object segmentation. These additional components enable robust object detection, accurate BBox prediction, and effective instance segmentation.

### 2.1. Experimental setup

The experimental setup consists of a motor, three sheaves (one sheave for holding weight, two rotation pulley blocks, two wire guide wheels), four LED lights, NVIDIA Jetson Nano P3450, and three defective SFRs each of length 8 m subjected to a weight of 50 kg (Rani et al., 2023). During the data collection process, the SFRs were rotated on sheaves supported by rotation pulleys and wheels for guiding the rope used for lifting purposes. This rotational movement helps simulate real-world scenarios where the ropes are subjected to rotational forces. The motor was operated at approximately 6 rotations per second to prevent motion blur when capturing images at 165 fps with an exposure time of 0.01 seconds. Although the back-and-forth movement of the SFRs over the sheaves may potentially cause faults to shift or redistribute along the rope's surface, no significant changes were observed in fault distribution during the experimental process. This suggests that the artificially introduced faults remained relatively stable throughout the experiments. The experimental setup is shown in Fig. 2.

### 2.2. Rope description

Dyneema (Dynamica-ropes aps) is a gel-spun, multi-filament fibre manufactured from HMPE (high-modulus-poly-ethene) or UHMWPE (ultra-high-molecular-weight-poly-ethylene). It possesses several notable characteristics such as high strength, low weight, low elongation at break, and resistance to most chemicals or harsh environments. These excellent mechanical properties with low density, result in a high performance-to-weight ratio. It serves as a valuable resource for researchers and industry professionals to monitor and assess the conditions of fibre ropes as potential replacements for SWRs. In the present work, Dyneema fibre is used to construct the rope for conducting the experiment. The characteristics of the rope used in the experiment are given as follows:

- Fibers: Dyneema SK 75/78
- Nominal Diameter: 8 mm
- 12 strands / 12 braided rope
- Torsional neutral
- Pitch/stitch length approx. 11mm.
- Braiding period: approx. 66mm

### 2.3. Data collection

The image dataset has been acquired using a Basler acA2000 camera with a Basler C11-5020-12M-P Premium 12-megapixel lens. To read the images from the camera, an NVIDIA Jetson Nano P3450 was used as the processing (for reading and analyzing the captured data) platform. To provide sufficient and uniform illumination, four Aputure AL-MC RGBWW LED lights each having an illumination of 1000 lux have been used. A total of 1,803 images having a resolution of 2000 × 1080 pixels have been collected to apply the defect detection algorithms. Each rope used in the present consists of 20 placking, 6 compression, and 4 chafing defects. This dataset comprises multiple instances of the same errors or faults occurring in the rope at different time intervals, along with images capturing various types of defects observed in different sections of the rope. Defects such as compression, which involve changes in diameter over an extended length of the rope, cannot be considered as single defects due to their elongated nature. Importantly, all images in the dataset are original, and no data augmentation techniques were applied. Each image offers a unique representation of SFRs, showcasing diverse defects and conditions without any artificial modifications introduced through augmentation methods.

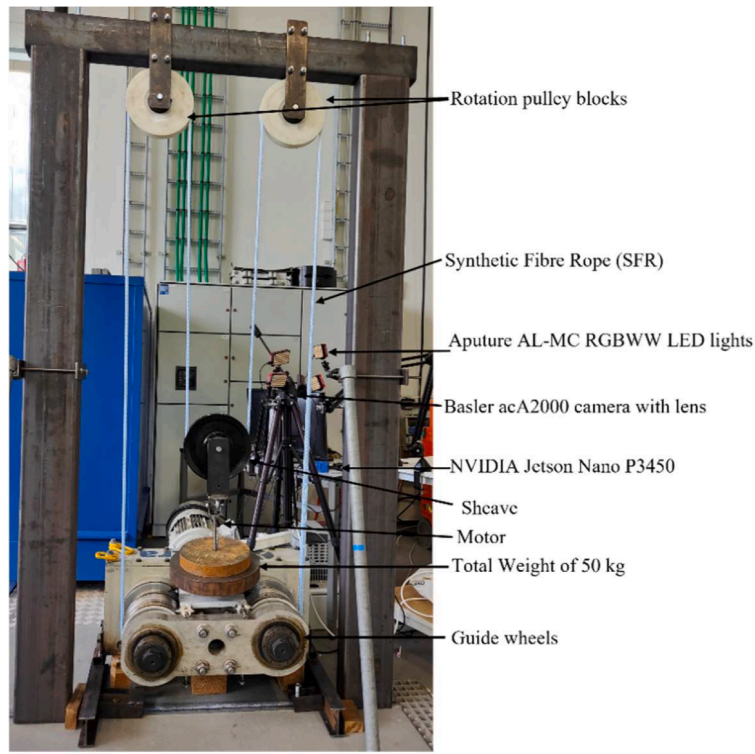

**Fig. 2.** Experimental setup for collecting SFRs dataset.

The ISO standard 9554:2019 "Fibre rope – General specifications" provides comprehensive information regarding the potential damages that may occur during the lifespan of SFRs (Iso 9554 2019). This standard serves as a valuable reference for understanding the types of defects that can occur in these ropes (Lian et al., 2017, Lin et al., 2022, Li et al., 2023, Davies et al., 2015). The experiment has been performed on three SFRs where defects have been artificially introduced by an expert roper from Dynamica Aps, Denmark (Dynamica-ropes aps n.d.). Fig. 3 depicts the various rope damages considered in the paper. The three ropes contain placking defects (pulled strands) as one of the potential defects that can occur in various applications. The severity of the placking defects varied, ranging from high to medium levels. In addition to the placking defects, each of the three defective ropes also contained other types of defects, including compression, chafing, and core out. By including these specific defects in the experiment, the aim was to provide a realistic representation of the types and severity levels of defects that can occur in SFRs during their operational lifespan. The dataset used in this study (utilizes three ropes) is a subset of the complete image dataset (utilizes ten ropes) that is publicly accessible and can be found in Rani et al. (2023). The detailed layout of the collected SFRs dataset related to the number of images in each set, defect type, and distribution is depicted in Table 1.

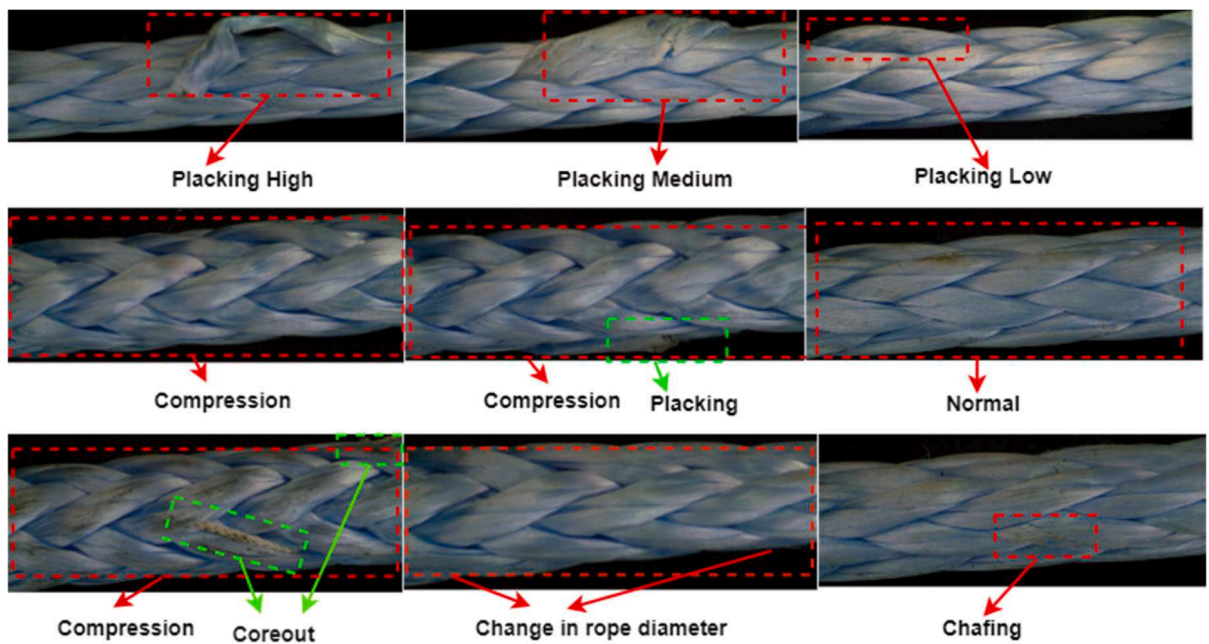

**Fig. 3.** Examples of data collected for rope with various damages and normal.

*2.4. Object detection and segmentation*

Detectron2 has robust object detection and segmentation capabilities. It can accurately identify and differentiate objects even when they overlap or occlude each other. This is especially important in scenes with multiple objects of varying sizes and orientations. As a result, Detectron2 provides rich information beyond classification, allowing for spatial understanding, precise localization, and instance level analysis.

The primary outcome of an object detection method is in the form of BBox. In the case of SFRs, the defects to be detected are generally asymmetric in shape. Therefore, a rectangular BBox may not be suitable for annotation. In such cases, polygons can be used as an alternative solution for annotating defects. A polygon may have arbitrary points, making it more accurate for defining the defects in SFRs. In the present work, labeling was performed on the collected dataset with the polygon annotation method using VGG image annotator (VIA), an open-source tool. The labeled dataset includes seven classes: placking high, placking medium, placking low, compression, core out, chafing and normal respectively. Fig. 4 depicts the annotated image using the VGG annotator.

*2.5. Model configuration*

The training configuration is described as follows:

- Training dataset: A custom train dataset of SFRs was introduced to the platform consisting of 1,315 images each having a dimension of 2040 × 1086 pixels.
- Learning rate (LR): The learning rate for the model was set to 0.00025.
- Total iterations: The training phase was performed for a total of 30,000 iterations.
- Number of classes: The model was configured to have 7 classes: placking high, placking medium, placking low, compression, core out, chafing and normal respectively. These classes represent various damages on the SFRs.
- Threshold: The testing threshold for object detection was set to 0.70. During the inference, objects with a detection score above 0.70 are considered positive detections.

**Table 1**
Layout of collected SFRs dataset.

| Defect Type | Number of Images |
|---|---|
| Placking Low | 120 |
| Placking Medium | 121 |
| Placking High | 403 |
| Compression | 827 |
| Core out | 278 |
| Chafing | 54 |
| Total Images | 1,803 |

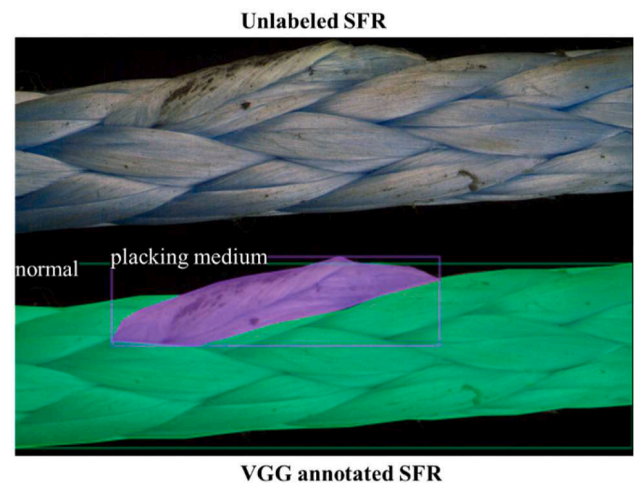

**Fig. 4.** Annotated sample using VGG annotator.

## 3. Results

The Detectron2 framework, commonly used for both object detection and segmentation tasks involving generic objects, was utilized with the Mask R-CNN architecture to detect defects in an experimentally collected dataset of SFRs. It was implemented on a system featuring a CPU speed of 3 GHz, 32 GB RAM, and an Nvidia GeForce RTX 4090 GPU. The entire dataset of 1,803 images has been divided into three subsets: train, validation, and test images. The model was trained with 1,315 images and then fine-tuned and validated over 331 images. Also, 157 images have been used for evaluating the model's performance, serving as the test images. To address the dataset's limited size and prevent overfitting, various data augmentation techniques were applied, including rotation, flipping, and adjustments in brightness and contrast. However, it's important to note that Detectron2's performance can be influenced by the nature of the objects being detected and the dataset characteristics. Comparisons between generic and non-generic objects, such as industrial defects, may not be directly applicable due to potential disparities in object distributions and model generalization.

*3.1. Performance comparison of backbone architectures*

Initially, the paper compares the performance of Mask R-CNN with ResNet-50-FPN-3x, ResNet-101-FPN-3x, and ResNeXt-101-FPN-3x architectures respectively. The model's nomenclature follows a format: [backbone]-[feature]-[learning schedule], where the backbone represents the chosen DL neural network architecture, and the feature denotes the feature extraction methodology using FPN. The training was performed using TL by initializing the model's weights by pre-training it on the COCO dataset for 3 epochs (notated as 3x learning schedule). The architectures are chosen based on the high average precision (AP) on object segmentation BBox's and instance segmentation masks. AP is commonly used in object detection tasks, particularly when there is a need to prioritize precision over other metrics like recall, or when dealing with imbalanced datasets. Given that the dataset includes defects of different sizes that need to be detected, AP has been chosen as the performance metric to evaluate the model's ability to accurately detect objects of varying sizes. Including AP in the evaluation allows for a comprehensive assessment of the model's precision-recall trade-off across different object sizes, which is essential for understanding its performance in object detection tasks with diverse object sizes. AP is calculated by computing the area under the precision-recall curve. It summarizes the model's performance across all confidence thresholds. AP ranges from 0 to 1, where higher values indicate better performance. AP values in the range of 0.7 to 0.9 are considered very good for defect detection tasks. Table 2 presents the performance comparison of Mask R-

**Table 2**
Comparison of the performance metrics of pre-trained models obtained from the training dataset.

| Model | Train time (s/iter) | | Type | AP$^{50}$(%) | |
| --- | --- | --- | --- | --- | --- |
| | Pre-trained | Custom-dataset | | Pre-trained | Custom-dataset |
| ResNet-50-FPN-3x | 0.26 | **0.13** | Box | 41.0 | **77.01** |
| | | | Segm | 37.2 | **77.97** |
| ResNet-101-FPN-3x | 0.34 | 0.16 | Box | 42.9 | 66.66 |
| | | | Segm | 38.6 | 66.66 |
| ResNext-101-FPN-3x | 0.69 | 0.22 | Box | **44.3** | 66.41 |
| | | | Segm | **39.5** | 66.20 |

CNN on the pre-trained dataset and custom dataset. The IoU evaluation metric has been used to measure the accuracy of the object detector given by Eq. (1). Here, the result of IoU greater than 0.5 (AP$^{50}$) is considered a good forecast. It can be observed that ResNeXt-101-FPN-3x shows high AP$^{50}$ for both BBox (44.3 %) and segmentation mask (39.5 %) with a comparatively large training time of 0.690 s/iter in comparison to other backbone models. Though ResNeXt-101-FPN has better AP$^{50}$ for BBox and mask on but takes longer time to train or predict. However, in the case of the custom dataset results show high AP$^{50}$ for BBox (77.01 %) and segmentation (77.97 %) in less training time of 0.1267 s/iter for ResNet-50-FPN-3x. Therefore, Mask R-CNN with ResNet-50-FPN-3x is chosen as the backbone architecture for the present work. Fig. 5 depicts the AP for BBox and instance segmentation for ResNet-50-FPN-3x architecture for the experimentally collected dataset of SFRs.

Also, detailed performance metrics of different backbone architectures for different IoUs and object sizes have been obtained on the

**Table 3**
Comparison of the performance metrics of different models obtained from the training dataset in order to choose the best-suited model for in-situ application.

| Model | Type | AP (%) | AP$^{50}$(%) | AP$^{75}$(%) | AP$^m$(%) | AP$^l$(%) |
| --- | --- | --- | --- | --- | --- | --- |
| ResNet-50-FPN-3x | Box | **58.37** | **77.01** | **64.95** | 23.30 | **59.40** |
| | Segm | **58.57** | **77.97** | **65.51** | 24.10 | **59.67** |
| ResNet-101-FPN-3x | Box | 53.88 | 66.66 | 59.19 | **25.74** | 54.80 |
| | Segm | 51.82 | 66.66 | 57.25 | **25.39** | 52.92 |
| ResNext-101-FPN-3x | Box | 55.56 | 66.42 | 62.79 | 25.49 | 56.60 |
| | Segm | 52.19 | 66.19 | 58.13 | 25.32 | 53.68 |

* Image size is 512 × 512.
* AP is average precision, AP$^{50}$ and AP$^{75}$ is the AP computed at an IoU value of 0.50 and 0.75 respectively.
* AP$^m$ and AP$^l$ are AP for medium and large objects.

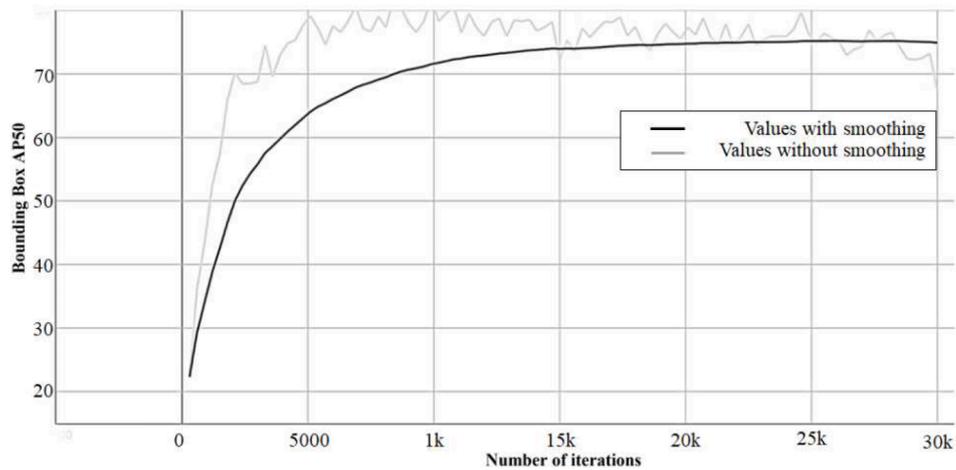

(a) AP at IoU=0.50 for BBox.

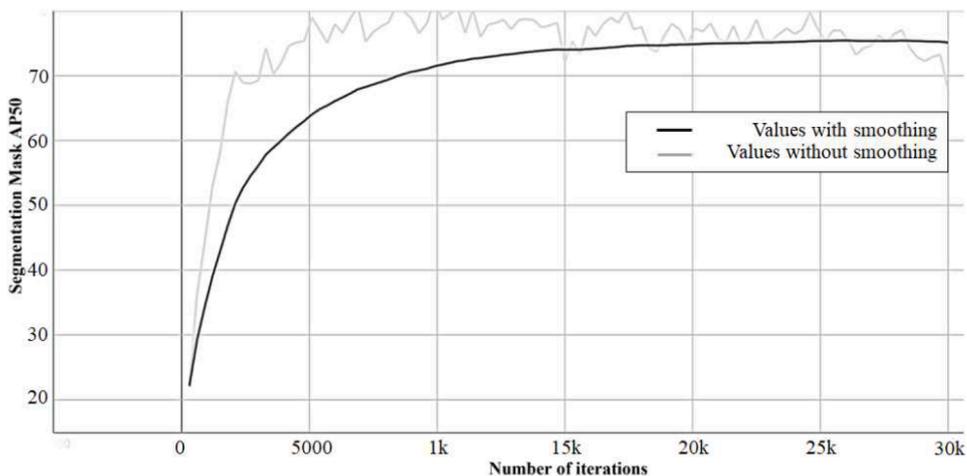

(b) AP at IoU=0.50 for instance segmentation.

**Fig. 5.** Evaluation results for ResNet-50-FPN

custom dataset of SFRs. Table 3 presents the performance metrics for ResNet-50-FPN, ResNet-101-FPN and ResNeXt-101-FPN. In each case, the training was performed for 30,000 iterations with a base learning rate of 0.00025. The disparity between the best values for $AP^m$ and $AP^l$ highlights the model's sensitivity to object size variations. The model may perform significantly better in detecting larger objects compared to medium-sized ones. Further, the difference in these AP values also provides insights into the distribution of object sizes within the dataset. It can be observed during the validation of backbone architectures, that the Mask R-CNN with ResNet-50-FPN backbone presented the best results for the precision metrics in less training time distributed for BBox and mask type in comparison to ResNet-101-FPN and ResNeXt-101-FPN backbone architectures. While the present work focuses on evaluating the precision of defect detection based on object size (medium and large), it is crucial to consider that precision can also vary significantly across different defect types in SFRs. The size and shape of defects can exhibit substantial diversity across types, irrespective of their severity levels. For instance, placking defects are larger while core out defects could appear smaller and more localized. This inherent variation in size and shape, independent of defect type or severity, can impact the model's ability to precisely detect and localize different defects. Furthermore, defect types with high representation in the training dataset are likely to be detected more precisely compared to those with limited samples, owing to the model's exposure during training. Additionally, certain defect types, such as core out involving internal breakages, may be more challenging to detect and segment precisely when compared to surface-level defects like compression or chafing, affecting class-wise precision.

*3.2. Data augmentation*

The training dataset encompasses a considerable number of defective images and defect types. However, despite this diversity, an inherent imbalance among the datasets may still exist. To address this, various data augmentation techniques on the collected SFR dataset have been applied in the present work. These techniques include resizing, horizontal/vertical flipping, rotation, and random adjustments to contrast and brightness. The augmentation was applied prior to training the model, serving the purpose of augmenting the instances of damages within the dataset. For resizing, the dataset was transformed with min and max sizes set at 800 pixels. The brightness and contrast augmentation were applied with a random probability of 0.5 with a maximum limit of 0.2. Similarly, vertical flipping was applied with a random probability of 0.5. In the present work, horizontal flipping does not have any impact on the current dataset.

Additionally, rotation was incorporated within a range of ±15 degrees on the collected dataset. The model was subsequently trained with these augmented datasets. Table 4 provides an overview of the performance metrics for the chosen Mask R-CNN with R50-FPN-3x architecture for augmented and non-augmented SFRs datasets. The results indicate that the performance of the ResNet-50-FPN architecture with annotations is notably inferior when compared to the non-annotated model, resulting in $AP^{50}$ scores of 50.15 % and 49.52 % for object detector bounding boxes and instance segmentation masks, respectively.

**Table 4**
Comparison of the performance metrics obtained from the training dataset for chosen Mask R-CNN ResNet-50-FPN-3x architecture with and without augmentation.

| Mask R-CNN-R50-FPN-3x | Type | AP (%) | $AP^{50}$(%) | $AP^{75}$(%) | $AP^m$(%) | $AP^l$(%) |
|---|---|---|---|---|---|---|
| Without augmentation | Box | **58.37** | **77.01** | **64.95** | 23.30 | **59.40** |
|  | Segm | **58.57** | **77.97** | **65.51** | 24.10 | **59.67** |
| With augmentation | Box | 27.66 | 50.15 | 28.55 | **23.57** | 28.41 |
|  | Segm | 27.51 | 49.52 | 27.27 | **23.60** | 28.46 |

* Image size is 512 × 512.

The brightness and contrast technique did not yield any noticeable improvement in the performance. This suggests that the collected dataset already encompasses images with different light and weather conditions. Also, the collected dataset contains images with different orientations due to the rotation of ropes across the sheaves while collecting the dataset. Therefore, rotation techniques do not have any impact on the performance of the model. Similarly, other augmentation techniques, including flipping and resizing, did not contribute to enhancing the model's performance. In conclusion, neither of these augmentation approaches appeared to improve the model's overall performance.

*3.3. Visual evaluation*

Fig. 6 presents the segmentation mask and accuracy of the polygon applied to each instance in SFRs for the Mask R-CNN with ResNet-50-FPN architecture. The predicted polygon also has corresponding labels and a confidence percentage ranging from 0 to 100 % as low to high confidence of having a defect on the SFR. This confidence parameter (threshold value) provides an estimation for the detected defect to be a true positive (TP). In this case, the threshold value is set to 0.70. If the set value is very low, then the results may show a large number of non-secure detections. The results show that model has successfully segmented each defect instance, providing precise boundaries and outlines for the detected defects. This instance segmentation capability is particularly beneficial when defects overlap or occlude each other, as it allows for accurate separation and identification of individual defect instances.

*3.4. Model assessment*

The performance of the model is illustrated in Fig. 7.

1. Accuracy: The model was trained with 1,315 images and achieved a training accuracy of approximately 97.6% as shown in Fig. 7 (a). The accuracy represents the percentage of correct predictions made by the model compared to the actual data. During testing, the trained model was evaluated using a separate test dataset consisting of 157 images. Out of the 157 test images, 154 were correctly identified as defective by the model with a high confidence score threshold of over 0.70 (threshold). Despite the complexity and variability of images, the model maintains a high level of accuracy in detecting defects in SFRs.
2. False Positive (FP) curve: Fig. 7 (b) shows the FP curve, which represents the instances where the model identified a defect incorrectly (FP) as the training progressed. It can be observed that the FP curve decreases as the training progresses, indicating the model's improved ability to correctly detect defects with a high confidence score.
3. False Negative (FN) curve: Fig. 7 (c) illustrates the false negative curve, which represents instances where the model failed to identify a defect (FN) with a confidence score below 0.70. The FN curve decreased as the training steps increased, indicating that the model improved in minimizing false negatives and accurately identifying defects with high confidence. A lower false negative rate is crucial for ensuring that critical defects are not missed during inspections. Despite the high performance of the model, as demonstrated in the Fig. 7, there are instances where the model may struggle with accurate defect detection. Fig. 8 illustrates few misinterpreted or incorrectly classified defects obtained from the model. In the top image, the model was not able to detect 'core out' defect correctly due to the small and low contrast of the defect. While in the bottom image, the SFR is not a completely compressed rope rather a normal rope leading to failed inspection.
4. Loss: The total loss value and validation loss, depicted in Fig. 7 (d) and (e), indicates the number of mistakes made by the model during each training or validation iteration. It can be observed that the total loss reduces to 0.18 after 30,000 iterations, indicating the model's improved performance over time.

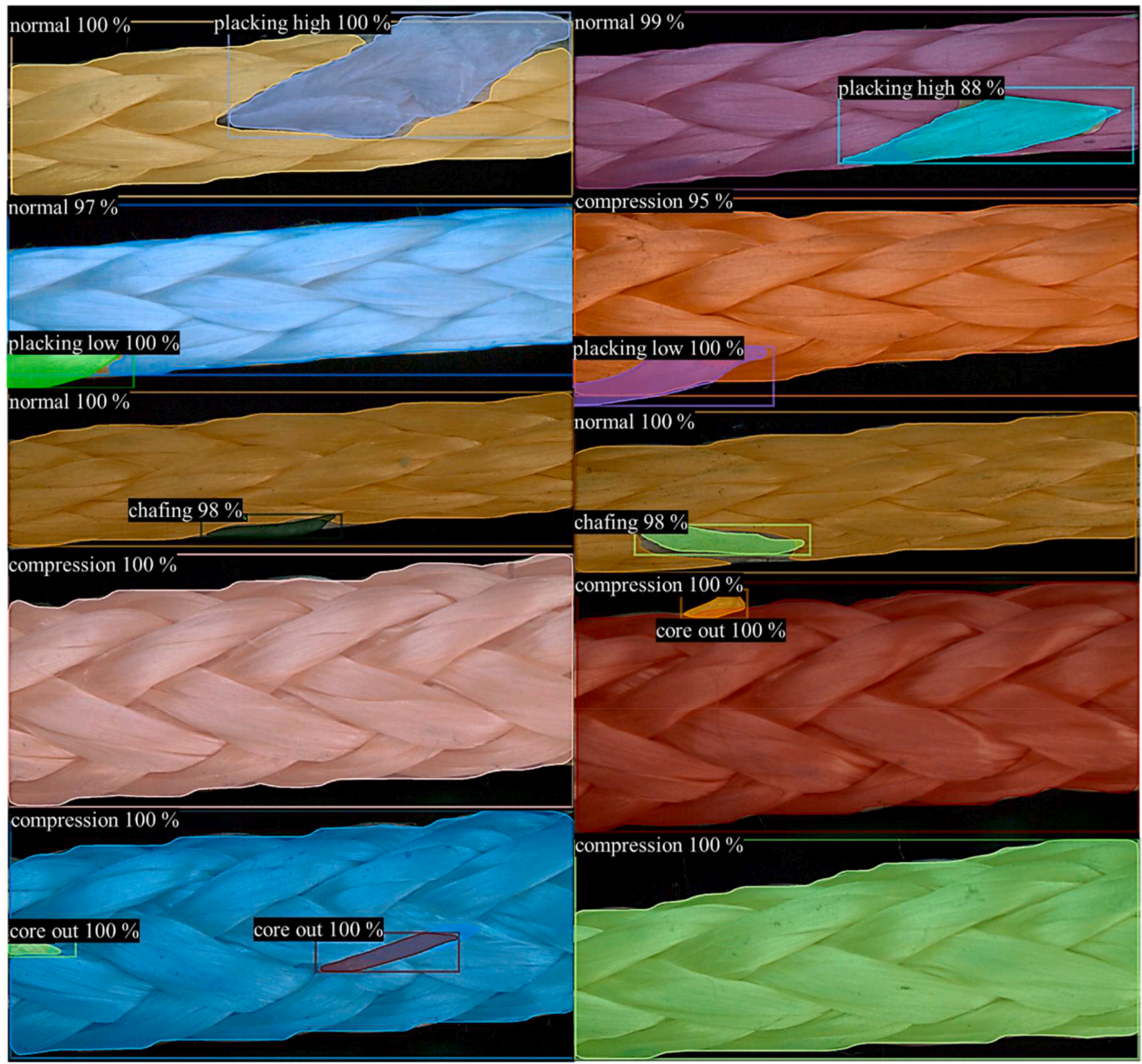

**Fig. 6.** Output results obtained from Detectron2.
* The color is randomly chosen to enhance the visual clarity and make it easier for readers to distinguish between different defects or classes detected by the model.

Overall, the provided results demonstrate the effectiveness of Detectron2 framework with Mask R-CNN in accurately detecting defects in SFRs, particularly those with curved, irregular boundaries or overlapping defects (e.g., placking, chafing, and compressions). This information can be leveraged for more targeted maintenance and repair strategies, as well as for developing a deeper understanding of the failure mechanisms and degradation patterns specific to SFRs. The high accuracy achieved, along with the observed improvements in the FP and FN curves, suggests the model's potential for practical application in defect detection tasks for SFRs. With appropriate fine-tuning and adaptation, Detectron2 could potentially be extended to inspect and monitor the condition of ropes made from different materials, further expanding its applicability and impact. Nonetheless, the model faces challenges when dealing with smaller, less prominent, highly occluded/blended, or low-contrast defects (e.g., core outs). Moreover, uncertainties may arise during the integration of Detectron2 for SFR defect detection, encompassing aspects like convergence issues, overfitting, or underfitting during training, as well as the optimal selection of hyperparameters and evaluation metrics. Addressing these uncertainties and optimizing model parameters are vital steps in augmenting the model's efficacy and dependability in practical settings.

While the present paper focuses on defect detection at a specific point in time, the model's precise defect characterization could facilitate monitoring defect progression over time. By analyzing sequential images or video data of the same rope section, the model's ability to accurately locate and segment defects could allow for monitoring the progression of defects, such as the growth of plackings, chafings, or the expansion of compression zones. This information could be valuable for predictive maintenance and RUL estimation of SFRs.

### 4. Conclusion and future work

The paper presents a state-of-the-art Detectron2 framework that uses Mask R-CNN with ResNet50-FPN architecture for object detection and instance segmentation. We conducted a comparative analysis involving ResNet101-FPN and ResNeXt101-FPN architectures alongside

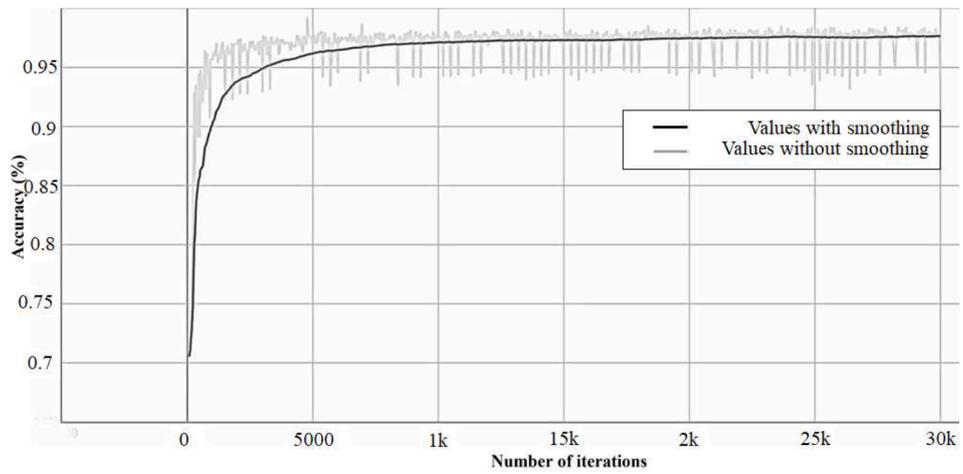

(a)   Mask R-CNN Accuracy

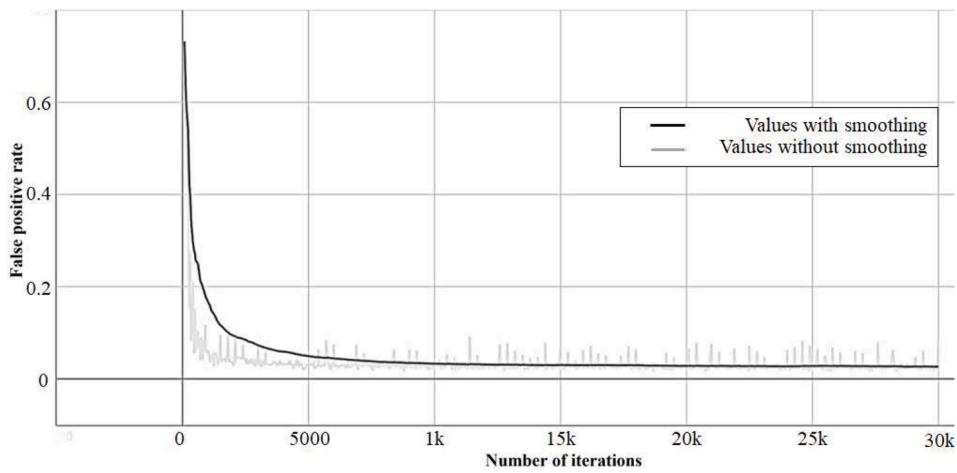

(b)   False Positives while training

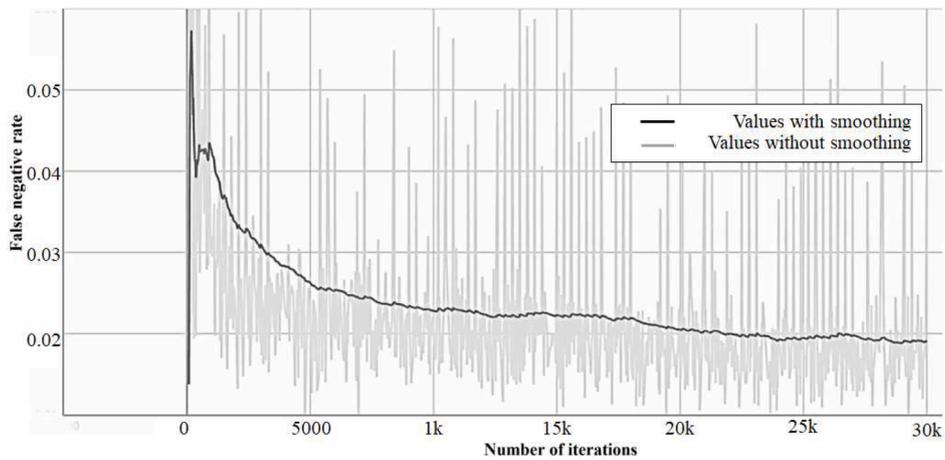

(c)   False Negatives while training

**Fig. 7.** Evaluation results for the proposed model

ResNet50-FPN to determine the most suitable model for in-situ applications. The effectiveness of the model was evaluated on an experimentally collected dataset obtained from introducing artificial damages on real SFRs. The model's performance demonstrates its precision in detecting damages across diverse categories, including those with irregular and complex shapes. The model highlights its potential for practical application in damage detection tasks for SFRs. During the model's training, a key challenge arises from the need for manual annotation to evaluate and understand the performance of each backbone architecture. Another constraint emerges due to the limited number of experimentally collected SFRs images for training the model. To enhance the model's performance, synthetic data may be incorporated into the dataset. This approach is anticipated to contribute substantially to elevating the model's effectiveness in condition monitoring applications. Moreover, it's crucial to recognize the potential complexity of real-world scenarios such environmental conditions (e.g., temperature, humidity), variations in rope manufacturing processes, and operational parameters (e.g., tension, stress), where multiple failure types may occur simultaneously or sequentially over the rope's lifetime. While the present model is trained on six damage types obtained experimentally in controlled environments to minimize external influences and variability. It may struggle to provide accurate descriptions when ropes exhibit combinations of these damage types not adequately represented in the training dataset. To address this limitation and improve the model's ability to handle complex failure scenarios, several strategies can be explored. These include expanding the training dataset to include a more diverse range of damage types and combinations, incorporating multi-label classification techniques to predict multiple damage types simultaneously, and exploring advanced deep-learning models capable of capturing complex interactions between different failure modes. In the future, this study could be expanded to include further experiments or simulations involving various types of ropes, including SWRs, SFRs, or hybrid ropes, and exploring different materials and constructions. Through a systematic comparison of the methodology's performance across diverse rope types, materials, and constructions, researchers can evaluate its effectiveness and pinpoint any limitations or challenges it may encounter.

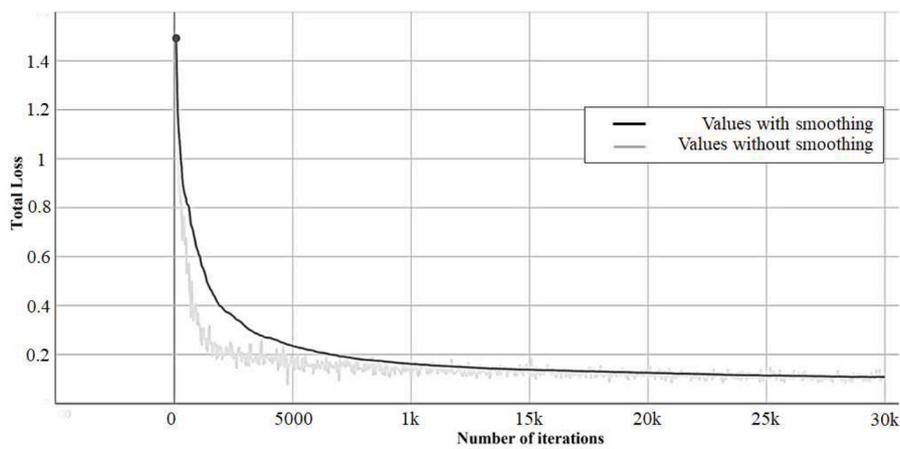

(d) Total Loss

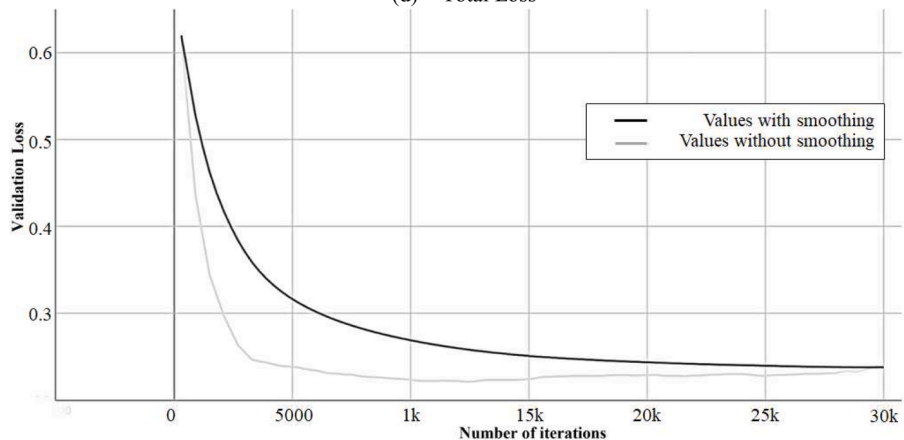

(e) Validation Loss

**Fig. 7.** (*continued*).

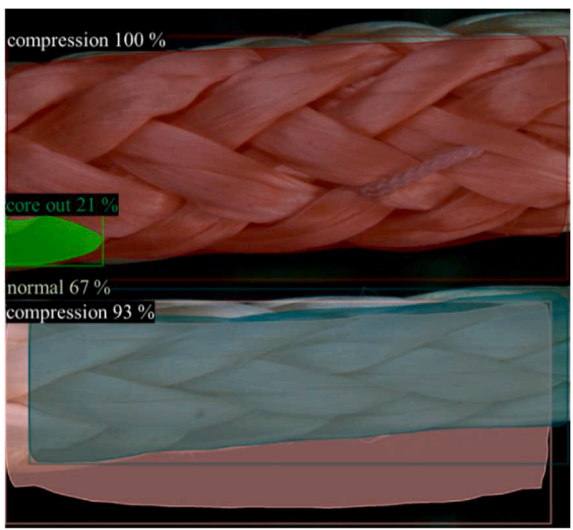

**Fig. 8.** Illustration of few misinterpreted damages obtained from the model.

## CRediT authorship contribution statement

**Anju Rani:** Writing – review & editing, Writing – original draft, Visualization, Software, Methodology, Investigation, Formal analysis, Data curation, Conceptualization. **Daniel Ortiz-Arroyo:** Writing – review & editing, Supervision, Project administration, Funding acquisition, Conceptualization. **Petar Durdevic:** Writing – review & editing, Supervision, Project administration, Funding acquisition, Data curation, Conceptualization.

## Declaration of competing interest

The authors declare that they have no known competing financial interests or personal relationships that could have appeared to influence the work reported in this paper.

## Data availability

No data was used for the research described in the article.

## Acknowledgements

This research was supported by Aalborg University, Liftra ApS (Liftra), and Dynamica Ropes ApS (Dynamica) in Denmark under the Energiteknologiske Udviklings- og Demonstrationsprogram (EUDP) program through project grant number 64021-2048.

## References

Ali, A.A., Katta, R., Jasek, R., Chramco, B., Krayem, S., 2022. COVID-19 detection from chest X-ray images using Detectron2 and Faster R-CNN. In: Proceedings of the Computational Methods in Systems and Software. Cham. Springer International Publishing, pp. 37–53.

Antin, K-N., Machado, M.A., Santos, T.G., Vilaça, P., 2019. Evaluation of different non-destructive testing methods to detect imperfections in unidirectional carbon fiber composite ropes. Journal of Nondestructive Evaluation 38, 1–12.

Bharati, P., Pramanik, A., 2020. Deep learning techniques—r-cnn to mask r-cnn: a survey. In: Computational Intelligence in Pattern Recognition: Proceedings of CIPR 2019, pp. 657–668.

Bolya, D., Zhou, C., Xiao, F., Lee, Y.J., 2019. Yolact: Real-time instance segmentation. In: Proceedings of the IEEE/CVF international conference on computer vision, pp. 9157–9166.

Cai, Z., Vasconcelos, N., 2018. Cascade r-cnn: Delving into high quality object detection. In: Proceedings of the IEEE conference on computer vision and pattern recognition, pp. 6154–6162.

Casey, N.F., Taylor, J.L., 1985. The evaluation of wire ropes by acoustic emission techniques. British Journal of Non-Destructive Testing 27 (6), 351–356.

Chen, Liang-Chieh, Zhu, Yukun, Papandreou, George, Schroff, Florian, Adam, Hartwig, 2018. Encoder-decoder with atrous separable convolution for semantic image segmentation. In: Proceedings of the European conference on computer vision (ECCV), pp. 801–818.

Davies, P., François, M., Lacotte, T.D., Vu, N., Durville, D., 2015. An empirical model to predict the lifetime of braided hmpe handling ropes under cyclic bend over sheave (cbos) loading. Ocean Engineering 97, 74–81.

Dynamica-ropes aps, Denmark.

Falconer, S., Gromsrud, A., Oland, E., Grasmo, G., 2017. Preliminary results on condition monitoring of fiber ropes using automatic width and discrete length measurements. In: Annual Conference of the PHM Society, 9.

Falconer, S., Nordgård-Hansen, E., Grasmo, G., 2020. Computer vision and thermal monitoring of hmpe fibre rope condition during cbos testing. Applied Ocean Research 102, 102248.

Feyrer, K., 2007. Wire ropes, 317. Springer-Verlag Berlin Heidelberg, Berlin, p. 2007.

Girshick, R., 2015. Fast r-cnn. In: Proceedings of the IEEE international conference on computer vision, pp. 1440–1448.

Halabi, Yahia, Xu, Hu, Yu, Zhixiang, Alhaddad, Wael, Dreier, Isabelle, 2023. Experimental-based statistical models for the tensile characterization of synthetic fiber ropes: a machine learning approach. Scientific Reports 13 (1), 17768.

He, K., Gkioxari, G., Dollár, P., Girshick, R., 2017. Mask r-cnn. In: Proceedings of the IEEE international conference on computer vision, pp. 2961–2969.

Hoppe, L.F.E., 1997. Performance improvement of dyneema (r) in ropes. In: Oceans' 97. MTS/IEEE Conference Proceedings, 1. IEEE, pp. 314–318.

Huang, X., Liu, Z., Zhang, X., Kang, J., Zhang, M., Guo, Y., 2020. Surface damage detection for steel wire ropes using deep learning and computer vision techniques. Measurement 161, 107843.

Iso 9554 (2019). fibre ropes – general specifications.

Kirillov, A., He, K., Girshick, R., Rother, C., Dollár, P., 2019. Panoptic segmentation. In: Proceedings of the IEEE/CVF conference on computer vision and pattern recognition, pp. 9404–9413.

Li, W., Li, G.and, Ye, H., Li, H., Ge, Y., Lin, S., 2023. Experimental study on cyclic-bend-over-sheave (cbos) characteristics of an hmpe fibre rope under dynamic loading. Applied Ocean Research 138, 103642.

Lian, Y., Liu, H., Zhang, Y., Li, L., 2017. An experimental investigation on fatigue behaviors of hmpe ropes. Ocean Engineering 139, 237–249.

Lin, S., Li, G., Li, W., Ye, H., Li, H., Pan, R., Sun, Y., 2022. Experimental measurement for dynamic tension fatigue characteristics of hmpe fibre ropes. Applied Ocean Research 119, 103021.

Liu, W., Anguelov, D., Erhan, D., Szegedy, C., Reed, S., Fu, C-Y., Berg, A.C., 2016. Ssd: Single shot multibox detector. In: Computer Vision–ECCV 2016: 14th European Conference, Amsterdam, The Netherlands, October 11–14, 2016, Proceedings, Part I 14. Springer, pp. 21–37.

Long, J., Shelhamer, E., Darrell, T., 2015. Fully convolutional networks for semantic segmentation. In: Proceedings of the IEEE conference on computer vision and pattern recognition, pp. 3431–3440.

McKenna, H.A., Hearle, J.W.S., O'Hear, N., 2004. Handbook of fibre rope technology, 34. Woodhead publishing.

Oland, E., Schlanbusch, R., Falconer, S., 2017. Condition monitoring technologies for synthetic fiber ropes-a review. International Journal of Prognostics and Health Management 8 (2).

Onur, Y.A., İmrak, C.E., 2011. The influence of rotation speed on the bending fatigue lifetime of steel wire ropes. Proceedings of the Institution of Mechanical Engineers, Part C: Journal of Mechanical Engineering Science 225 (3), 520–525.

Paixao, J., da Silva, S., Figueiredo, E., Radu, L., Park, G., 2021. Delamination area quantification in composite structures using gaussian process regression and auto-regressive models. Journal of Vibration and Control 27 (23-24), 2778–2792.

Pham, V., Pham, C., Dang, T., 2020. Road damage detection and classification with detectron2 and faster r-cnn. In: 2020 IEEE International Conference on Big Data (Big Data). IEEE, pp. 5592–5601.

Platzer, E-S., Süße, H., Nägele, J., Wehking, K-H., Denzler, J., 2010. On the suitability of different features for anomaly detection in wire ropes. In: Computer Vision, Imaging and Computer Graphics. Theory and Applications: International Joint Conference, VISIGRAPP 2009gall, February 5-8, 2009. Revised Selected Papers. Lisboa, Portu. Springer, pp. 296–308.

A. Rani, D. O. Arroyo, and P. Durdevic. Imagery dataset for condition monitoring of synthetic fibre ropes. arXiv preprint arXiv:2309.17058, 2023.

Ren, S., He, K., Girshick, R., Sun, J., 2015. Faster r-cnn: Towards real-time object detection with region proposal networks. Advances in neural information processing systems 28.

Ridge, I.M.L., Chaplin, C.R., Zheng, J., 2001. Effect of degradation and impaired quality on wire rope bending over sheave fatigue endurance. Engineering Failure Analysis 8 (2), 173–187.

Ronneberger, Olaf, Fischer, Philipp, Brox, Thomas, 2015. U-net: Convolutional networks for biomedical image segmentation. In: Medical Image Computing and Computer-Assisted Intervention–MICCAI 2015: 18th International Conference, Munich, Germany, October 5-9, 2015, Proceedings, Part III 18. Springer, pp. 234–241.

Ke Sun, Yang Zhao, Borui Jiang, Tianheng Cheng, Bin Xiao, Dong Liu, Yadong Mu, Xinggang Wang, Wenyu Liu, and Jingdong Wang. High-resolution representations for labeling pixels and regions. arXiv preprint arXiv:1904.04514, 2019.

Tabernik, D., Šela, S., Skvarč, J., Skočaj, D., 2020. Segmentation-based deep-learning approach for surface-defect detection. Journal of Intelligent Manufacturing 31 (3), 759–776.

Tuomas Jalonen, Mohammad Al-Sa'd, Roope Mellanen, Serkan Kiranyaz, and Moncef Gabbouj. Real-time damage detection in fiber lifting ropes using convolutional neural networks. arXiv preprint arXiv:2302.11947, 2023.

Vallan, A., Molinari, F., 2009. A vision-based technique for lay length measurement of metallic wire ropes. IEEE Transactions on Instrumentation and Measurement 58 (5), 1756–1762.

Wang, X., Shrivastava, A., Gupta, A., 2017. A-fast-rcnn: Hard positive generation via adversary for object detection. In: Proceedings of the IEEE conference on computer vision and pattern recognition, pp. 2606–2615.

Wei, X., Yang, Z., Liu, Y., Wei, D., Jia, L., Li, Y., 2019. Railway track fastener defect detection based on image processing and deep learning techniques: A comparative study. Engineering Applications of Artificial Intelligence 80, 66–81.

Weller, SD, Davies, Peter, Vickers, AW, Johanning, Lars, 2015. Synthetic rope responses in the context of load history: The influence of aging. Ocean Engineering 96, 192–204.

Wen, H., Huang, C., Guo, S., 2021. The application of convolutional neural networks (CNNs) to recognize defects in 3D-printed parts. Materials 14 (10), 2575.

Wu Y., Kirillov A., Massa W.-Y., Lo F., and Girshick R., Detectron 2, 2019.

Yagüe F. J., Diez-Pastor J. F., Latorre-Carmona P., and Osorio C. I. G., Defect detection and segmentation in X-Ray images of magnesium alloy castings using the Detectron2 framework. arXiv preprint arXiv:2202.13945, 2022.

Yan, X., Zhang, D., Pan, S., Zhang, E., Gao, W., 2017. Online nondestructive testing for fine steel wire rope in electromagnetic interference environment. NDT & E International 92, 75–81.

Yang, J., Li, S., Wang, Z., Yang, G., 2019. Real-time tiny part defect detection system in manufacturing using deep learning. IEEE Access 7, 89278–89291.

Yang, J., Li, S., Wang, Z., Dong, J., Wang, H., Tang, S., 2020. Using deep learning to detect defects in manufacturing: a comprehensive survey and current challenges. Materials 13 (24), 5755.

Ye, Haoran, Li, Wenhua, Lin, Shanying, Ge, Yangyuan, Lv, Qingtao, 2024. A framework for fault detection method selection of oceanographic multi-layer winch fibre rope arrangement. Measurement, 114168.
Zhang, D.L., Cao, Y.N., Wang, C., Xu, D.G., 2006. A new method of defects identification for wire rope based on three-dimensional magnetic flux leakage. In: Journal of Physics: Conference Series, 48. IOP Publishing, p. 334.
Zhao, Hengshuang, Shi, Jianping, Qi, Xiaojuan, Wang, Xiaogang, Jia, Jiaya, 2017. Pyramid scene parsing network. In: Proceedings of the IEEE conference on computer vision and pattern recognition, pp. 2881–2890.
Zhu, X., Lyu, S., Wang, X., Zhao, Q., 2021. Tph-yolov5: Improved yolov5 based on transformer prediction head for object detection on drone-captured scenarios. In: Proceedings of the IEEE/CVF international conference on computer vision, pp. 2778–2788.

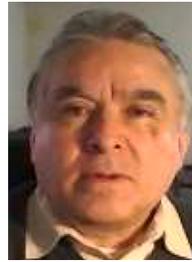

**Dr. Daniel Ortiz-Arroyo** is currently an associate professor in the Department of Energy at Aalborg University in Denmark. He earned a Ph.D. in Computer Science and Engineering at Oregon State University, USA. He has edited several books and is the author of more than 70 papers in international journals and conferences. His areas of research are in AI and Machine Learning, Robotics, Fuzzy Logic, NLP, security, computer architecture, and social network analysis.

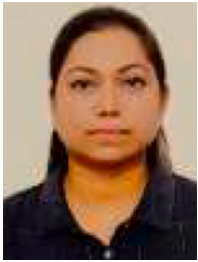

**Dr. Anju Rani** received an M. Tech from Thapar University, India and a Ph.D. from the Department of Electrical Engineering, Indian Institute of Technology Ropar, India. She is currently a post-doctorate in the Department of Energy at Aalborg University, Denmark. Her research interests include thermal non-invasive/non-destructive imaging technologies, condition monitoring, deep learning, and bio-medical image processing.

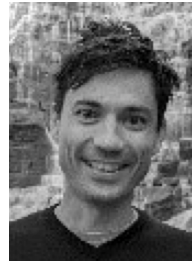

**Dr. Petar Durdevic** received a BSc in data and electronics and an MSc degree in control systems from Aalborg University, Denmark, in 2011 and 2013, respectively. He received his Ph. D. in Control Engineering from Aalborg University, Denmark in 2017. He is currently an associate professor at the Department of Energy Technology at Aalborg University, Denmark. His current research interests include nonlinear system identification and control, artificial intelligence, and condition monitoring with application domains in robotics and industrial processes.